\documentclass{article}

\usepackage[final]{nips_2018}

\usepackage{natbib}        % for arXiv submission (their natbib is of different version)

\usepackage[utf8]{inputenc} % allow utf-8 input
\usepackage[T1]{fontenc}    % use 8-bit T1 fonts
\usepackage{hyperref}       % hyperlinks
\usepackage{url}            % simple URL typesetting
\usepackage{booktabs}       % professional-quality tables
\usepackage{amsfonts}       % blackboard math symbols
\usepackage{nicefrac}       % compact symbols for 1/2, etc.
\usepackage{microtype}      % microtypography
\usepackage{multirow}
\usepackage{makecell}       % thick line
\usepackage{graphicx}
\usepackage{subcaption}
\usepackage{amsmath}
\usepackage{amssymb}
\usepackage{color,soul}

\usepackage{xcolor}
\usepackage[normalem]{ulem}
\def\JNdel#1{\bgroup\markoverwith{\textcolor{blue}{\rule[0.5ex]{2pt}{1pt}}}\ULon{#1}}

\title{Semi-supervised classification by reaching consensus among modalities}

\author{Zining Zhu$^{1,2}$, Jekaterina Novikova$^{1}$, Frank Rudzicz$^{3,5,2,4,1}$ \\
  $^1$: Winterlight Labs, Toronto ON, Canada \\
  $^2$: University of Toronto, ON, Canada \\
  $^3$: Li Ka Shing Knowledge Institute, St Michael's Hospital, Toronto ON, Canada\\
  $^4$: Vector Institute for Artificial Intelligence, Toronto ON, Canada \\
  $^5$: Surgical Safety Technologies, Toronto ON, Canada\\
  {\tt \{zining, jekaterina\}@winterlightlabs.com, frank@cs.toronto.edu} \\}

\begin{document}

\maketitle

\begin{abstract}
Deep learning has demonstrated abilities to learn complex structures, but they can be restricted by available data. Recently, Consensus Networks (CNs)\cite{CN2018} were proposed to alleviate data sparsity by utilizing features from multiple modalities, but they too have been limited by the size of labeled data. In this paper, we extend CN to Transductive Consensus Networks (TCNs), suitable for semi-supervised learning. In TCNs, different modalities of input are compressed into latent representations, which we encourage to become indistinguishable during iterative adversarial training. 
To understand TCNs two mechanisms, consensus and classification, we put forward its three variants in ablation studies on these mechanisms. To further investigate TCN models, we treat the latent representations as probability distributions and measure their similarities as the negative relative Jensen-Shannon divergences. We show that a consensus state beneficial for classification desires a stable but imperfect similarity between the representations.
Overall, TCNs outperform or align with the best benchmark algorithms given 20 to 200 labeled samples on the Bank Marketing and the DementiaBank datasets.
 
\end{abstract}

\section{Introduction}
Deep learning has demonstrated impressive capacities to learn complicated structures from massive data sets. However, acquiring sufficient labeled data can be expensive or difficult (e.g., for specific pathological populations \citep{lee2017medical}). 
Transductive learning (a set of semi-supervised algorithms) uses intrinsic structures among unlabeled data to boost classifier performance. 
In the real world, data can spread across multiple modalities (e.g., visual, acoustic, and text) in typical tasks, although many existing transductive algorithms do not exploit the structure across these modalities. Co-training \citep{blum1998co-training} and tri-training \citep{zhou2005tri} use one classifier per modality to supervise each other, but they can only apply to two and three modalities respectively.

Recently, Consensus Networks (CNs) \citep{CN2018} incorporated the idea of co-training. Not limited by the number of modalities, CNs showed promising results on detecting cognitive impairments from  multi-modal datasets of speech. A consensus network contains several interpreters (one per modality), a discriminator, and a classifier. The interpreters try to produce low-dimensional representations of input data that are indistinguishable by the discriminator. The classifier makes predictions based on these representation vectors.

Despite promising results, CN is limited by the amount of available training data. This motivates our extension into semi-supervised learning with our Transductive Consensus Network (TCN). 

TCNs operate in two mechanisms: as consensus or classifier. The consensus mechanism urges the modality representations to resemble each other (trained on the whole dataset without using labels), and the classifier mechanism optimizes the networks to retain information useful for classification (trained on the labeled dataset). 
To illustrate the importance of these two mechanisms in an ablation study, we also put forward its three variants: TCN-embed, TCN-svm, and TCN-AE in \S \ref{sec:tcn-model}. By this ablation study, we show that both mechanisms should function together via iterative training.

To further reveal the mechanisms of TCN, we formulate in \S \ref{sec:similarity} the similarity between latent representations using negative Jensen-Shannon divergences. By monitoring their similarities, we show that a meaningful consensus state prefers representations to have suboptimal similarities. 

In experiments (\S \ref{sec:experiments}), we compare TCN to its three variants, TCN's multimodal supervised learning counterpart (CN), and several other semi-supervised learning benchmark algorithms on two datasets: Bank Marketing (from the UCI repository) and DementiaBank (a dataset of pathological speech in multiple modalities). 
On both datasets, the F-scores of TCN align with the best benchmark models when there are more labeled data available, and outperform benchmarks (including tri-training) given as few as 20 labeled points.

\section{Related work}

Transductive SVMs \citep{joachims1999tsvm} were an early attempt in transductive semi-supervised learning. In addition to the SVM objective, TSVMs minimize the hinge loss on unlabeled data. TSVMs have yielded good performance on our datasets, so we include them for completeness.

Later, many semi-supervised learning algorithms took either autoencoding or GAN approaches.
In autoencoding, a model learns a low-dimensional representation and a reconstruction for each data sample. Usually, noise is added in generating the low-dimensional representation. By trying to minimize the difference between reconstructed and original data, the model learns (i) an encoder capturing low-dimensional hidden information and (ii) a decoder, which is a generative model able to recover data. This is the approach of the denoising autoencoder \citep{vincent2010stacked,lee2013pseudo}. An extension is Ladder network \citep{rasmus2015ladder}, which stacks denoising autoencoders and adds layer-wise reconstruction losses. Ladder networks are often more computationally efficient than stacked denoise autoencoders. %In recent years, variational autoencoders are also used for semi-supervised learning \citep{kingma2014semi,maaloe2016auxiliary}, by maximizing the variational lower bound of the log likelihood probability of data.\TODO{Add ADGM as benchmark if mentioned here}

In generative adversarial networks (GANs) \citep{goodfellow2014generative}, a generator tries to produce data that are indistinguishable from true data, given a discriminator which itself learns to tell them apart. This adversarial training procedure could proceed with few labeled data points. For example, Feature-matching GANs \citep{salimans2016improved} add generated (``synthetic'') samples into the training data of the discriminator as an additional class. Another example is Categorical GANs \citep{springenberg2016categoricalGAN} which optimize uncertainty (measured by entropy of predictions) in the absence of labels. Noticeably, \citet{dai2017good} showed that a discriminator performing well on a training set might not benefit the whole dataset. CNs and TCNs, despite not containing generative components, are built with the adversarial principles inspired from GANs.

The idea to make multiple components in the network agree with each other has been adopted by several previous models. For example,  \citet{benediktsson1997parallel} proposed Parallel Consensus Networks, where multiple networks classify by majority voting. Each of the networks is trained on features after a unique transform. \citet{mescheder2017numerics} proposed consensus optimization in GAN in which a term is added to the utility functions of both the generator and discriminator to alleviate the adversity between generator and discriminator. However, neither they nor semi-supervised learning utilized the multiple modalities.

Multi-modal learning is also referred to as multi-view learning. \cite{Pou-Prom2018} computed multiple viewpoints from speech samples and classified cognitive impairments. By contrast, our multi-view learning is semi-supervised, and can involve non-overlapping subsets of features. 

In domain adaptation, some work has been applied to find a unified representation between domains, for example, by applying domain invariant training \cite{Ganin2016} and semantic similarity loss \cite{Motiian2017}. However, our approach does not involve multiple domains -- we only handle data from one domain. Here, the term `domain' refers to how the data are naturally generated, whereas the term `modality' refers to how different aspects of data are observed.

\section{Methodology}
\label{sec:tcn-model}
In previous work, Consensus Networks \citep{CN2018} was proposed for multimodal supervised learning. We extend the model to be suitable for semi-supervised learning, resulting in Transductive Consensus Networks (TCNs). This section also presents three variants: TCN-embed, TCN-svm, and TCN-AE. 

\subsection{Problem setting}
Given labeled data, $\{ \mathbf{x^{(i)}}, y^{(i)}\}$ ($\mathbf{x^{(i)}} \in \mathcal{X_L}$), and unlabeled data, \{$\mathbf{x^{(i)}}$\} (where $\mathbf{x^{(i)}} \in \mathcal{X_U}$), we want to learn a model that reaches high accuracies in predicting labels in unlabeled data. In the semi-supervised learning setting, there are many more unlabeled data points than labeled: $|\mathcal{X_U}| \gg |\mathcal{X_L}|$, where the whole dataset is $\mathcal{X} = \mathcal{X_L} \cup \mathcal{X_U}$.

Each data point $\mathbf{x}$ contains feature values from multiple modalities (i.e., `views'). If $M$ be the total number of modalities, then $\mathbf{x}=[\mathbf{x_1}, \mathbf{x_2}, ..., \mathbf{x_M}]$. For $m=1, ..., M$, $\mathbf{x_m}$ could have different dimensions, but the dimension of $\mathbf{x_m^{(i)}}$ is consistent throughout the dataset. E.g., there may be 200 (100) acoustic (semantic) features for each data point.

\subsection{Consensus Networks}

\begin{figure}
\begin{subfigure}{.33\textwidth}
\includegraphics[scale=0.6]{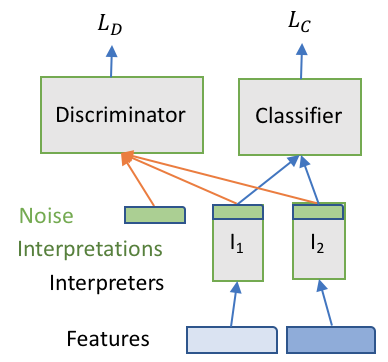}
\caption{TCN-vanilla and TCN-embed}
\end{subfigure}
\begin{subfigure}{.32\textwidth}
\includegraphics[scale=0.6]{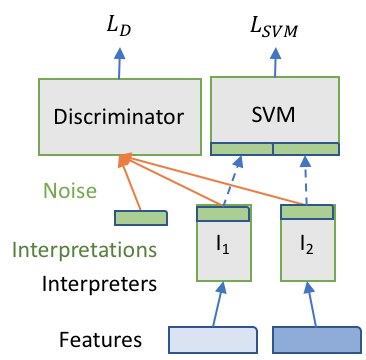}
\caption{TCN-svm}
\end{subfigure}
\begin{subfigure}{.32\textwidth}
\includegraphics[scale=0.55]{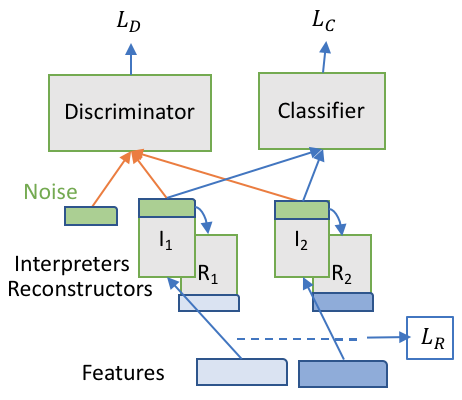}
\caption{TCN-AE}
\end{subfigure}
\label{fig:models}
\caption{Network structures for TCN and TCN-embed (left), TCN-svm (middle) and TCN-AE (right), taking the example when there are two modalities in data (M=2).
}
\end{figure}
Here we briefly review the structures of CNs. In a CN model, $M$ interpreter networks
$I_{1,...,M}(.)$ each compress the corresponding modality for a data sample into a low-dimensional vector $\mathbf{v}$. 
\[\mathbf{v_m} = I_m(\mathbf{x_m})\]
We call these networks {\em interpreters}, because they interpret the feature spaces with representations.
In TCNs, a consensus is expected to be reached given representations from multiple views of the same data sample.
A discriminator network $D(.)$ tries to identify the originating modality of each representation vector. If we write the $m^{th}$ modality of the dataset as a set $\mathcal{M}_m$ of vectors, then the discriminator function $D(.)$ can be expressed as:
\[ D(\mathbf{v_m}) = P(\mathbf{x_m} \in \mathcal{M}_m|I_{1..M},D) \]

To prevent the discriminator from looking at only superficial aspects for each data sample in the forward pass, Consensus Networks \cite{CN2018} include an additional `noise modality' representation sampled from a normal distribution, with mean and variance determined by the `non-noise' representations:
\[ \mathbf{v_{M+1}} \sim \mathcal{N}(\mathbf{\mu_{v_{1..M}}, \mathbf{\sigma^2_{v_{1..M}}}}) \]

The model's \textbf{discrimination loss} $\mathcal{L_{D}}$ is therefore defined as the cross entropy loss across all modalities (plus the noise modality), averaged across both labeled and unlabeled datasets $\mathcal{X}$:
\[ \mathcal{L}_{D} = \mathbb{E}_{\mathbf{x}} \mathbb{E}_{m} \Big \{-\text{log} P(\mathbf{x_m} \in \mathcal{M}_m|I_{1..M}, D) \Big \} \]

Finally, a classifier network $C(.)$ predicts the probability of class assignment ($y$) from the combined representation vectors, given model parameters: 
\[ C([\mathbf{v_1}, \mathbf{v_2}, ..., \mathbf{v_M}]) = P(y | \mathbf{x}, I_{1..M}, C)  \] 

The \textbf{classification loss} $\mathcal{L_{C}}$ is just the cross entropy loss across the labeled data:
\[ \mathcal{L_C} = \mathbb{E}_{x} \Big \{ - \text{log} P(y | \mathbf{x}, I_{1..M}, C) \Big \}\]

The \textbf{overall optimization goals for CN} can therefore be described as:
\begin{equation} \min_{C, I_{1..M}} \mathcal{L_C} \text{  and  } \max_{I_{1..M}}\min_{D} \mathcal{L_D} 
\label{eq:optim-goal-cn}
\end{equation}

\subsection{Transductive Consensus Network}
Consensus Networks (CNs), as a supervised learning framework, are limited by the amount of labeled data. This motivates us to generalize the approach to semi-supervised scenarios. There are two mechanisms in the CN training procedures, namely the {\em classifier} mechanism and {\em consensus} mechanism. The classifier mechanism requires labeled data, but the consensus mechanism does not explicitly require these labels. We let the consensus mechanism handle both labeled and unlabeled data. This results in Transductive Consensus Networks.

Formally, the loss functions are rewritten as:
\[ \mathcal{L}_{D} = \mathbb{E}_{\mathbf{x \in \mathcal{X}}} \mathbb{E}_{m} \Big \{-\text{log} P(\mathbf{x_m} \in \mathcal{M}_m|I_{1..M}, D) \Big \} \]

\[ \mathcal{L_C} = \mathbb{E}_{x\in \mathcal{X_L}} \Big \{ - \text{log} P(y | \mathbf{x}, I_{1..M}, C) \Big \}\]
where $\mathcal{X}$ consists of both labeled data $\mathcal{X_L}$ and unlabeled data $\mathcal{X_U}$. Overall, the optimization goal can still be written as:
\begin{equation} \min_{C, I_{1..M}} \mathcal{L_C} \text{  and  } \max_{I_{1..M}}\min_{D} \mathcal{L_D} 
\label{eq:optim-goal}
\end{equation}

These goals set up a complex nonlinear optimization problem. To figure out a solution, we break down the goals into three iterative steps, similar to GAN\cite{goodfellow2014generative}:
\begin{itemize} 
\item The `I' step encourages interpreters to produce indistinguishable representations: $\displaystyle \max_{I_{1..M}} \mathcal{L_D}$. 
\item The `D' step encourages discriminators to recognize modal-specific information retained in representations: $\displaystyle \min_{D} \mathcal{L_D}$. 
\item The `CI' step trains the networks to make a correct decision: $\displaystyle \min_{C, I_{1..M}} \mathcal{L_C}$.
\end{itemize}

\subsection{TCN variants}

The consensus mechanism builds a low-dimensional latent representation of each (labeled and unlabeled) data sample containing common knowledge across different modalities, and the classifier mechanism tries to make these representations meaningful. Three modifications are made to our base TCN model, resulting in the following models:

\paragraph{TCN-embed} consists of the same networks as TCN but is trained slightly differently. Before the I-D-CI optimization cycle, we add a pretraining phase with I-D iterations, which emphasizes the consensus mechanism.

\paragraph{TCN-svm} removes the classifier network from TCN-embed. After the pretraining phase across the whole dataset, we extract the representations of those labeled data samples to train a supervised learning classifier (i.e., an SVM). TCN-svm discards the classifier mechanism, which results in deteriorations to model performance (\S \ref{sec:results}).

\paragraph{TCN-AE} provides insights from another perspective.
In contrast to TCN, TCN-AE contains several additional reconstructor networks, $R_{1..M}(.)$ (one per modality). Each reconstructor network tries to recover the input modality from the corresponding low-dimensional representations (plus a small noise $\mathbf{\epsilon}$):
\[ \mathbf{\hat{x}_m} = R_m (\mathbf{v_m} + \mathbf{\epsilon}) \] 

Defining reconstruction loss as $\mathcal{L_R}=\displaystyle \mathbb{E}_{x \in \mathcal{X}} \mathbb{E}_{m} |\mathbf{\hat{x}_m} - \mathbf{x_m}|^2 $, the optimization target in TCN-AE can be expressed as:
\begin{equation} \min_{C, I_{1..M}, R_{1..M}} \mathcal{L_C} \text{, and } \max_{I_{1..M}}\min_{D} \mathcal{L_D} \text{, and } \min_{I_{1..M}, R_{1..M}} \mathcal{L_R}
\label{eq:optim-goal-AE}
\end{equation}
TCN-AE is inspired by denoising autoencoder\citep{vincent2010stacked}, where the existence of reconstructor networks encourage the latent variables to preserve realistic information. This somehow works against the consensus mechanism, which according to \cite{CN2018} tries to agree on simple representations. TCN-AE therefore weakens the consensus mechanism. We will show in \S \ref{sec:results} that an inhibited consensus mechanism results in inferior model performance.

\subsection{The similarity of representations}
\label{sec:similarity}
We want to quantitatively measure the effects of the consensus and the classification mechanisms.
To evaluate the similarities of representations, we treat the hidden dimensions of each representation $\mathbf{v_m}=[v_{m,1}, v_{m,2}, .., v_{m,j},...]$ (after normalization) as discrete values of a probability mass function\footnote{There is a ReLU layer at output of each interpreter network, so the probability mass will be non-negative.}, which we write as $p_m$. 
The $M$ modalities for each data point are therefore approximated by $M$ probability distributions. 
Now we can measure the relative JS divergences between each pair of representations $\mathbf{v_m}$ and $\mathbf{v_n}$ derived from the same data sample ($\hat{D}(p_m || p_n)$). 
To acquire the relative value, we normalize the JS divergence by the total entropy in $p_m$ and $p_n$:
\begin{gather*} 
\hat{D}(p_m || p_n) = \frac{1}{2(\mathbb{H}_{p_m} +  \mathbb{H}_{p_n})} (D_{KL}(p_m || p_n) + D_{KL}(p_n || p_m)) \\
\text{where } D_{KL}(p_m || p_n) = \sum_j v_{m,j} \text{log} \frac{v_{n,j}}{v_{m,j}}
\end{gather*}
where $v_{m,j}$ and $v_{n,j}$ are the $j^{th}$ component of $\mathbf{v_m}$ and $\mathbf{v_n}$ respectively.
In total, for each data sample with $M$ modalities, $\frac{M(M-1)}{2}$ pairs of relative divergences are calculated. We average the negative of these divergences to get the similarity:
\[ \displaystyle \text{Similarity} = \mathbb{E}_i \mathbb{E}_{m,n\in \{1..M\} \text{ and } m\neq n} \big \{-\hat{D}(p_m^{(i)} || p_n^{(i)}) \big \} \]

Note that by our definition the maximum value of the ``similarity'' value is 0 (where there is no JS divergence between any pair of the representation vectors), and it has no theoretical lower bound.

Figure \ref{fig:representations_tsne} shows several 2D visualizations of representation vectors drawn from an arbitrary run. In Figure \ref{fig:rmi_plot}, we illustrate how the similarities between modality representations evolve during training.

\section{Experiments}
\label{sec:experiments}
\subsection{Datasets}
We compare TCN and its variants on two benchmark datasets: Bank Marketing, and DementiaBank. The full list of features, by modalities, are provided in the Supplementary Material.

The Bank Marketing dataset is from the UCI machine learning repository\citep{Dua:UCI2017}.  
used for predicting whether the customer will subscribe a term deposit in a bank marketing campaign via telephone\citep{bank-marketing-dataset}. There are originally 4,640 positive samples (subscribed) and 36,548 negative ones (did not subscribe). Since consensus network models do not work well on imbalanced datasets, we randomly sample 5,000 negative samples to create an (almost) balanced dataset. We also convert the categorical raw features\footnote{\url{https://archive.ics.uci.edu/ml/datasets/bank+marketing}} into one-hot representations. We then divide the features into three modalities: basic information, statistical data, and employment-related features.

DementiaBank\footnote{\url{https://dementia.talkbank.org}} contains 473 spoken picture descriptions of the clinical ``cookie-theft picture''\citep{PittsCorpus}, containing 240 positive samples (the Dementia class) and 233 negative samples (the Control class). We extract 413 linguistic features from each speech sample and their transcriptions, including acoustic (e.g., pause durations), lexical \& semantic (e.g., average cosine similarities between words in sentences) and syntactic (e.g., complexity of the syntactic parse structures) modalities.

\begin{table}[h]
\centering 
\begin{tabular}{c c c c}
\Xhline{2\arrayrulewidth} 
Dataset & N. of samples & \%pos / \%neg & N. features per modality \\ \hline 
Bank marketing & 9640 & 48.13 / 51.87 & 10 / 22 / 12 \\
DementiaBank & 473 & 50.76 / 49.26 & 185 / 117 / 110 \\
\Xhline{2\arrayrulewidth}
\end{tabular}
\caption{Basic information about the datasets (after preprocessing). In the Bank Marketing dataset, the three modalities correspond to basic information, statistical data, and employment-related features. In DementiaBank, the three modalities correspond to acoustic, syntactic, and lexical\&semantic features. Detailed descriptions of the features are included in supplementary materials.}
\label{tab:datasets}
\end{table}

\subsection{Benchmarks}
We evaluate TCN and its variants against several benchmarks, including:
\begin{enumerate}
	\itemsep0em
    \item Multimodal semi-supervised learning benchmark: Tri-training \citep{zhou2005tri}\footnote{Implemented using three identical multiple layer perceptrons with one hidden layer containing 30 neurons.}.
	\item TCN's supervised counterpart: Consensus Network (CN).
    \item Unimodal semi-supervised learning: TSVM \citep{joachims1999tsvm}, ladder network \citep{rasmus2015ladder}, CatGAN \citep{springenberg2016categoricalGAN}.
\end{enumerate}

\subsection{Implementation}
\label{subsec:implementation}
For simplicity, we use fully connected networks for all of $I_{1..M}$, $D$, $C$, and $R_{1..M}$ in this paper. To enable faster convergence, all fully connected networks have a batch normalization layer \citep{ioffe2015batch}. For training, the batch size is set to 10. The neural network models are implemented using PyTorch \citep{pytorch}, and supervised learning benchmark algorithms (SVM, MLP) use scikit-learn \cite{scikit-learn}. 

We use the Adam optimizer \citep{kingma2014adam} with an initial learning rate of 0.001. In training TCN, TCN-embed, and TCN-AE, optimizations are stopped when the classification loss does not change by more than $10^{-5}$ in comparison to the previous step, or when the step count reaches 100. In the pre-training phase of TCN-embed and TCN-svm, training is stopped when the discrimination loss changes by less than $10^{-5}$, or when pretraining step count reaches 20. 

Sometimes, the iterative optimization (i.e., the I-D-CI cycle for TCN / TCN-embed, and the I-D-RI-CI cycle for the TCN-AE variant) is trapped in local saddle points -- the training classification loss does not change while the training classification loss is higher than $\log 2 \approx 0.693$. This is the expected loss of a binary classifier with zero knowledge. If the training classification loss is higher than $\log 2$, the model is re-initialized with a new random seed and the training is restarted. Empirically, this re-initialization happens no more than once per ten runs, but the underlying cause needs to be examined further. %\TODO{related to mode collapse?}

\section{Results and discussion}
\label{sec:results}
\begin{figure}[h]
    \includegraphics[width=.42\textwidth]{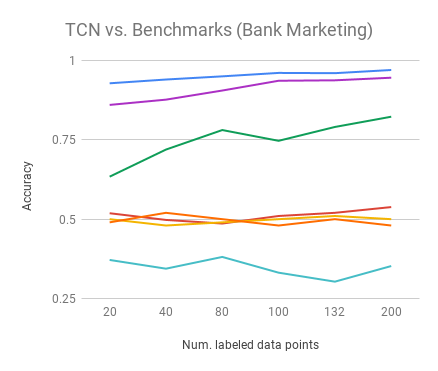}
    \includegraphics[width=.58\textwidth]{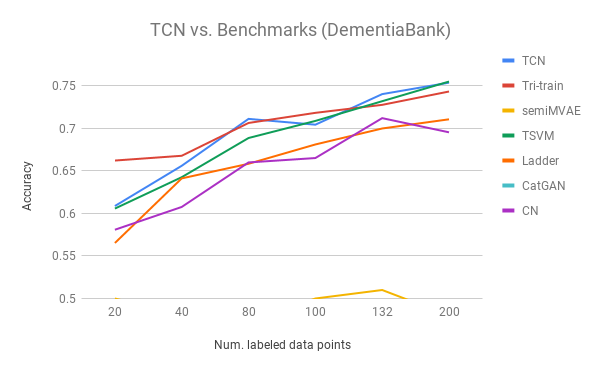}
    \caption{TCN (top blue lines; best viewed in colors) outperform or align with benchmark algorithms, including multi-modal semi-supervised (tri-train \cite{zhou2005tri}), uni-modal semi-supervised (TSVM \cite{joachims1999tsvm}, Ladder \cite{rasmus2015ladder}, CatGAN \cite{springenberg2016categoricalGAN}), and multi-modal supervised (CN \cite{CN2018}).}
    \label{fig:accuracy_tcn_vs_benchmarks}
\end{figure}

\begin{figure}[h]
    \includegraphics[width=.42\textwidth]{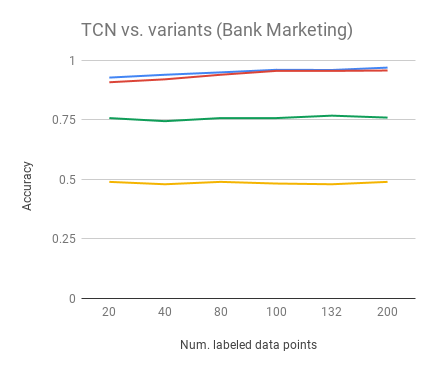}
    \includegraphics[width=.58\textwidth]{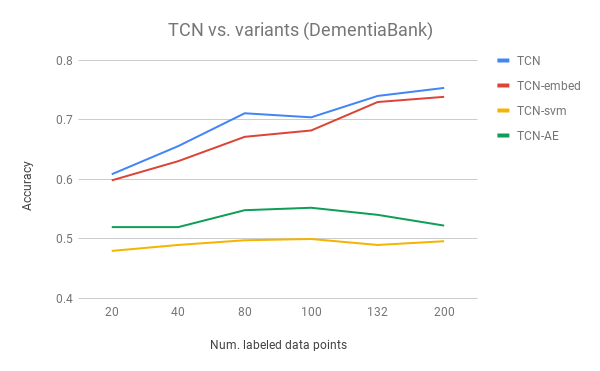}
    \caption{Accuracy plots for TCN vs its variants. Best viewed in colors. TCN-embed accuracies aligns with TCN, both significantly outperforming TCN-AE, which is better than TCN-svm. The consensus and classification mechanisms should both be present.}
    \label{fig:accuracy_tcn_vs_variants}
\end{figure}

\subsection{TCN performances versus the benchmarks}
As shown in Figure \ref{fig:accuracy_tcn_vs_benchmarks}, TCN outperforms or matches the best benchmarks. On the Bank Marketing dataset, TCN, CN, and TSVM clearly outperform the rest. On DementiaBank, Tri-train, TCN, and TSVM form the ``first tier''. 

Also, semi-supervised learning does not always outperform those supervised algorithms. For example, on the Bank Marketing dataset, CN (i.e., TCN's supervised learning counterpart) holds the second best performance.

% Moved forward for formatting
\begin{figure}[t]
\begin{subfigure}{0.32\textwidth}
\includegraphics[scale=0.205]{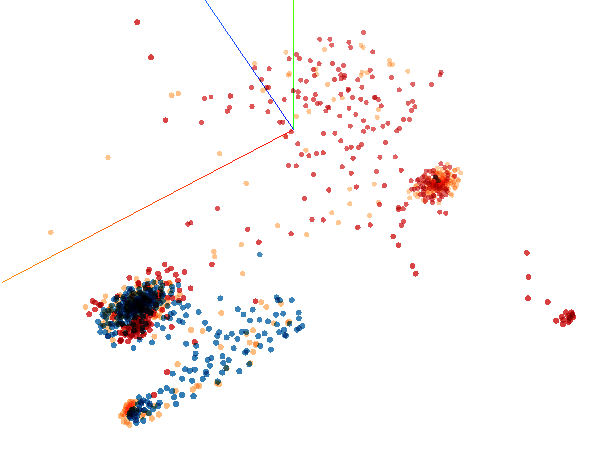}
\caption{Step 2: sim=-0.0206}
\end{subfigure}
\begin{subfigure}{0.32\textwidth}
\includegraphics[scale=0.165]{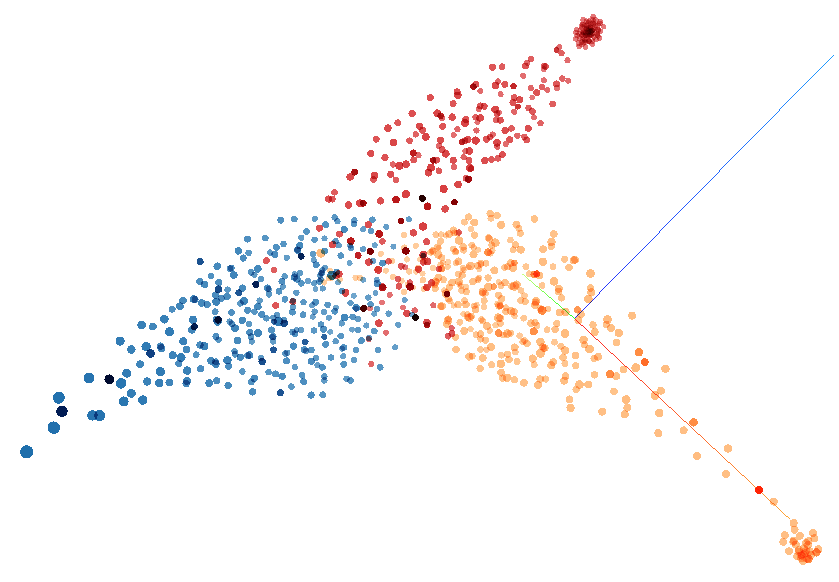}
\caption{Step 30: sim=-0.0072}
\end{subfigure}
\begin{subfigure}{0.32\textwidth}
\includegraphics[scale=0.07]{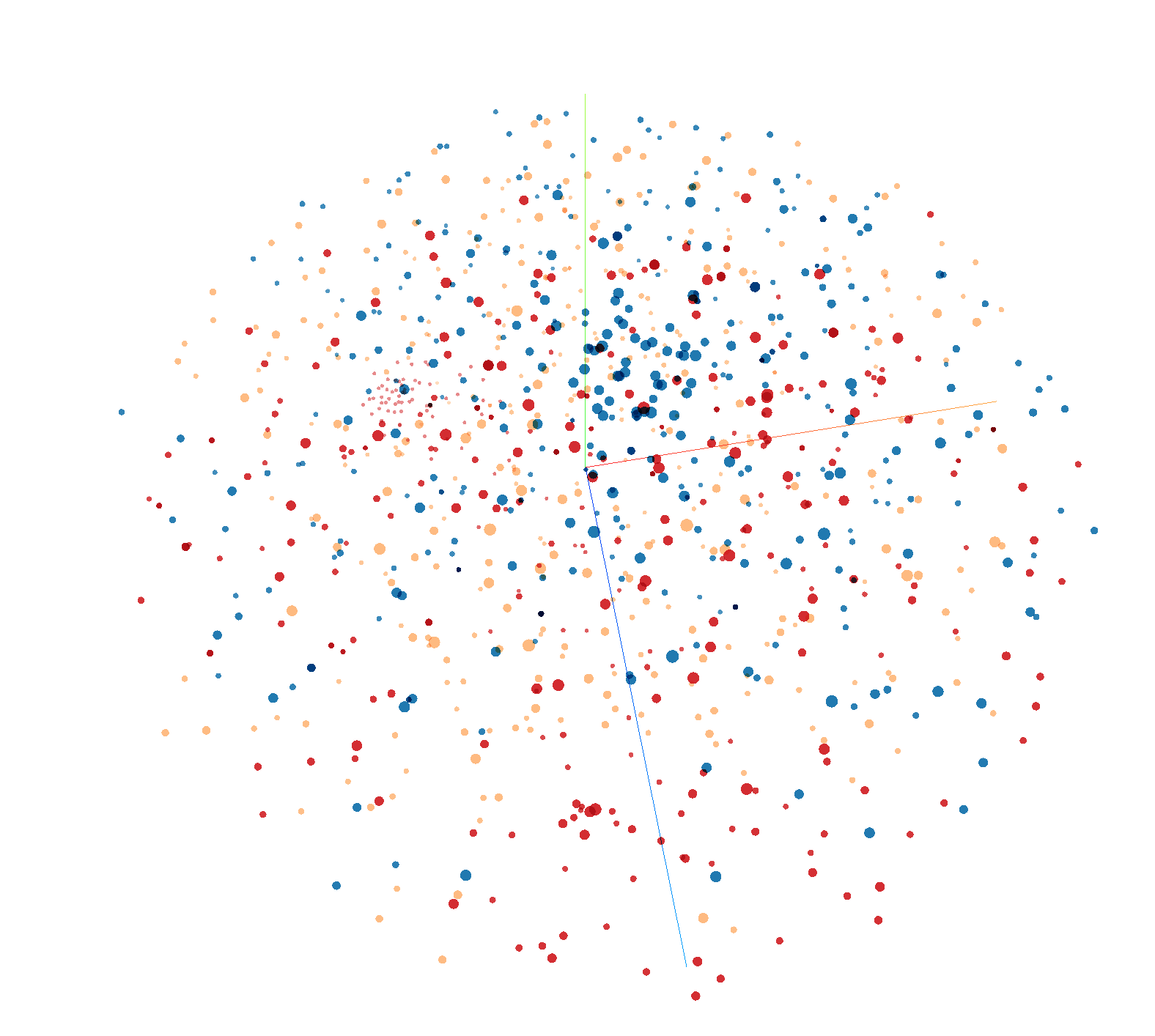}
\caption{Step 110: sim=-0.0146}
\end{subfigure}
\caption{Three 2-D T-SNE\citep{TSNE} visualizations of representations among modalities, taken from a run of the vanilla TCN (on DementiaBank dataset with 80 labeled data). The three colors represent three modalities. At step 2, the representations are distributed randomly. At step 110, they become mixed evenly. The most interesting embedding happens at step 30, when representations of the three modalities form three `drumstick' shapes. With the highest visual symmetry, this configuration also has the highest {\em similarity} among the three.
\label{fig:representations_tsne}}
\end{figure}

% Moved forward for formatting
\begin{figure}[t]
\centering 
\includegraphics[scale=.34]{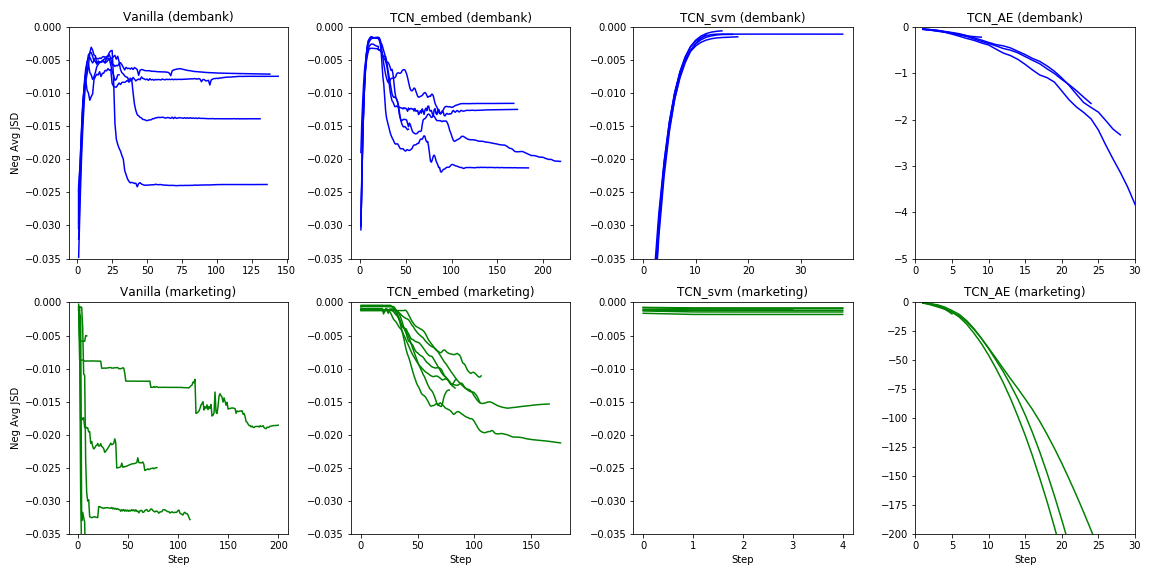} 
\caption{Examples of similarity plots against the number of steps taken, for DementiaBank using 80 labeled samples (``DB80'', blue) and Bank Marketing using 20 labeled samples (``BM20'', green). 
The $y$ axis are scaled to $(-0.035, 0)$ except TCN-AE, where the relative JS divergences ``explode''. Note that we stop the training procedure when losses converge (as detailed in \S \ref{subsec:implementation}), so the trials may stop at different steps.
\label{fig:rmi_plot}}
\end{figure}

\subsection{What can we learn from the TCN variants?}
\label{subsec:tcn-variants-eval}

As shown in Figure \ref{fig:accuracy_tcn_vs_variants}, TCN aligns with or outperforms TCN-embed. Both of these approaches significantly outperform TCN-AE. On the other hand, TCN-svm produces almost trivial classifiers. There are several points worth noting.
\begin{itemize}
    \item {\em Both} the consensus and the classification mechanisms are beneficial to classifier performance. The classification mechanism can be beneficial even with as few as 20 labeled data samples.
    \item Iterative optimization is crucial. Without classification mechanisms, the consensus mechanism by itself fails to derive good reprensentations. Without consensus mechanisms (i.e., when the reconstructors hinder the consensus mechanisms), accuracies drop significantly.
\end{itemize}

\subsection{Understanding the similarity of representations}
To understand more about TCN, we visualize them with T-SNE\cite{TSNE} in Figure \ref{fig:representations_tsne}, and plot the similarity values in Figure \ref{fig:rmi_plot}.
\begin{itemize}
    \item The higher similarity values corresponds to a state where the distributions contain higher symmetry in aggregate manner. 
    \item Measured from the similarity values, TCN models reach a consensus state where the similarities are stable. TCN-svm reaches agreements quickly but are close to trivial. TCN-AE, with the autoencoder blocking the consensus mechanism, fails to reach a state of agreement.
\end{itemize}

\section{Conclusion and future work}
In this paper, we present Transductive Consensus Networks (TCNs) that extend consensus networks with semi-supervised learning. We identify two mechanisms in which TCNs function, i.e., the consensus and classifier mechanisms. 
With three TCN variants in an ablation study, we show the importance of both mechanisms.
Moreover, by treating the representations as probability distributions and defining their similarity as negative relative JS divergences, we show that although the consensus mechanism urges high similarities, a good consensus state might not need perfect similarities between modality representations.

In the future, several avenues may be considered. To start with, building consensus networks using other types of neural networks may be considered. In addition, more exploration could be done to find a more explainable metric to describe the extent of agreement. Currently, we use $-\frac{D_{JS}}{\mathbb{H}_1 +\mathbb{H}_2}$, but this requires some approximations. Optimizing against the similarity metric, instead of setting up a discriminator, may be worth examining as well.

\bibliographystyle{plain}
\bibliography{bibliography}

\begin{thebibliography}{10}

\bibitem{PittsCorpus}
James~T Becker, Fran{\c{c}}ois Boiler, Oscar~L Lopez, Judith Saxton, and
  Karen~L McGonigle.
\newblock {The natural history of Alzheimer's disease: description of study
  cohort and accuracy of diagnosis}.
\newblock {\em Archives of Neurology}, 51(6):585--594, 1994.

\bibitem{benediktsson1997parallel}
Jon~Atli Benediktsson, Johannes~R Sveinsson, Okan~K Ersoy, and Philip~H Swain.
\newblock {Parallel consensual neural networks}.
\newblock {\em IEEE Transactions on Neural Networks}, 8(1):54--64, 1997.

\bibitem{blum1998co-training}
Avrim Blum and Tom Mitchell.
\newblock {Combining labeled and unlabeled data with co-training}.
\newblock In {\em Proceedings of the eleventh annual conference on
  Computational learning theory}, pages 92--100. ACM, 1998.

\bibitem{dai2017good}
Zihang Dai, Zhilin Yang, Fan Yang, William~W Cohen, and Ruslan~R Salakhutdinov.
\newblock {Good semi-supervised learning that requires a bad gan}.
\newblock In {\em Proceedings of Advances in Neural Information Processing
  Systems}, pages 6513--6523, 2017.

\bibitem{Dua:UCI2017}
Dua Dheeru and Efi {Karra Taniskidou}.
\newblock {{\{}UCI{\}} Machine Learning Repository}, 2017.

\bibitem{Ganin2016}
Yaroslav Ganin, Evgeniya Ustinova, Hana Ajakan, Pascal Germain, Hugo
  Larochelle, Fran{\c{c}}ois Laviolette, Mario Marchand, Victor Lempitsky, Urun
  Dogan, Marius Kloft, Francesco Orabona, and Tatiana Tommasi.
\newblock {Domain-Adversarial Training of Neural Networks}.
\newblock {\em Journal of Machine Learning Research}, 17:1--35, 2016.

\bibitem{goodfellow2014generative}
Ian Goodfellow, Jean Pouget-Abadie, Mehdi Mirza, Bing Xu, David Warde-Farley,
  Sherjil Ozair, Aaron Courville, and Yoshua Bengio.
\newblock {Generative Adversarial Nets}.
\newblock In {\em Proceedings of Advances in neural information processing
  systems}, pages 2672--2680, 2014.

\bibitem{ioffe2015batch}
Sergey Ioffe and Christian Szegedy.
\newblock {Batch normalization: Accelerating deep network training by reducing
  internal covariate shift}.
\newblock In {\em Proceedings of the 32nd International Conference on Machine
  Learning (ICML-15)}, pages 448--456, 2015.

\bibitem{joachims1999tsvm}
Thorsten Joachims.
\newblock {Transductive inference for text classification using support vector
  machines}.
\newblock In {\em Proceedings of the 16th International Conference of Machine
  Learning (ICML-99)}, volume~99, pages 200--209, 1999.

\bibitem{kingma2014adam}
Diederik Kingma and Jimmy Ba.
\newblock {Adam: A method for stochastic optimization}.
\newblock {\em Proceesings of International Conference on Learning
  Representations (ICLR)}, 2014.

\bibitem{lee2017medical}
Choong~Ho Lee and Hyung-Jin Yoon.
\newblock Medical big data: promise and challenges.
\newblock {\em Kidney research and clinical practice}, 36(1):3, 2017.

\bibitem{lee2013pseudo}
Dong-Hyun Lee.
\newblock {Pseudo-label: The simple and efficient semi-supervised learning
  method for deep neural networks}.
\newblock In {\em Workshop on Challenges in Representation Learning, ICML},
  volume~3, page~2, 2013.

\bibitem{TSNE}
Laurens van~der Maaten and Geoffrey Hinton.
\newblock {Visualizing data using t-SNE}.
\newblock {\em Journal of machine learning research}, 9(Nov):2579--2605, 2008.

\bibitem{mescheder2017numerics}
Lars Mescheder, Sebastian Nowozin, and Andreas Geiger.
\newblock The numerics of gans.
\newblock In {\em Proceesings of Advances in Neural Information Processing
  Systems}, pages 1823--1833, 2017.

\bibitem{bank-marketing-dataset}
S{\'{e}}rgio Moro, Paulo Cortez, and Paulo Rita.
\newblock {A data-driven approach to predict the success of bank
  telemarketing}.
\newblock {\em Decision Support Systems}, 62:22--31, 2014.

\bibitem{Motiian2017}
Saeid Motiian, Marco Piccirilli, Donald~A. Adjeroh, and Gianfranco Doretto.
\newblock {Unified Deep Supervised Domain Adaptation and Generalization}.
\newblock In {\em International Conference on Computer Vision (ICCV)}, 2017.

\bibitem{pytorch}
Adam Paszke, Sam Gross, Soumith Chintala, Gregory Chanan, Edward Yang, Zachary
  DeVito, Zeming Lin, Alban Desmaison, Luca Antiga, and Adam Lerer.
\newblock {Automatic differentiation in PyTorch}.
\newblock 2017.

\bibitem{scikit-learn}
F~Pedregosa, G~Varoquaux, A~Gramfort, V~Michel, B~Thirion, O~Grisel, M~Blondel,
  P~Prettenhofer, R~Weiss, V~Dubourg, J~Vanderplas, A~Passos, D~Cournapeau,
  M~Brucher, M~Perrot, and E~Duchesnay.
\newblock {Scikit-learn: Machine Learning in Python}.
\newblock {\em Journal of Machine Learning Research}, 12:2825--2830, 2011.

\bibitem{Pou-Prom2018}
Chlo{\'{e}} Pou-Prom and Frank Rudzicz.
\newblock {Learning multiview embeddings for assessing dementia}.
\newblock In {\em Empirical Methods in Natural Language Processing (EMNLP)2},
  2018.

\bibitem{rasmus2015ladder}
Antti Rasmus, Mathias Berglund, Mikko Honkala, Harri Valpola, and Tapani Raiko.
\newblock {Semi-supervised learning with ladder networks}.
\newblock In {\em Proceedings of Advances in Neural Information Processing
  Systems}, pages 3546--3554, 2015.

\bibitem{salimans2016improved}
Tim Salimans, Ian Goodfellow, Wojciech Zaremba, Vicki Cheung, Alec Radford, and
  Xi~Chen.
\newblock {Improved techniques for training gans}.
\newblock In {\em Proceedings of Advances in Neural Information Processing
  Systems}, pages 2234--2242, 2016.

\bibitem{springenberg2016categoricalGAN}
Jost~Tobias Springenberg.
\newblock {Unsupervised and semi-supervised learning with categorical
  generative adversarial networks}.
\newblock {\em Proceedings of International Conference on Learning
  Representations (ICLR)}, 2016.

\bibitem{vincent2010stacked}
Pascal Vincent, Hugo Larochelle, Isabelle Lajoie, Yoshua Bengio, and
  Pierre-Antoine Manzagol.
\newblock {Stacked denoising autoencoders: Learning useful representations in a
  deep network with a local denoising criterion}.
\newblock {\em Journal of Machine Learning Research}, 11(Dec):3371--3408, 2010.

\bibitem{zhou2005tri}
Zhi-Hua Zhou and Ming Li.
\newblock {Tri-training: Exploiting unlabeled data using three classifiers}.
\newblock {\em IEEE Transactions on knowledge and Data Engineering},
  17(11):1529--1541, 2005.

\bibitem{CN2018}
Zining Zhu, Jekaterina Novikova, and Frank Rudzicz.
\newblock {Detecting cognitive impairments by agreeing on interpretations on
  linguistic features}.
\newblock {\em arxiv 1808.06570}, 2018.

\end{thebibliography}

% \end{document}
\section*{Supplementary Material}

\subsection*{Features by modalities for Bank Marketing}
Three modalities are determined as following. The division are somewhat arbitrary, except that we try to make the binary features resulting from one categorical feature be in the same modality.
\begin{enumerate}
	\item Basic information: age, marital, education, housing, loan, contact, duration, pdays, previous, management
    \item Statistical information: campaign, poutcome, emp.var.rate, unknown, cons.conf.idx, euribor3m, day\_in\_week (converted to 7 binary features), month (converted to 12 binary features)
    \item Employment-related information: consumer price index, never employed, retired, self-employed, technician, services, student, housemaid, entrepreneur, blue-collar
\end{enumerate}

\subsection*{Features by modalities for Dementia Bank}
Three modalities are determined as following:
\begin{enumerate}
	\item Acoustic-related features: phonation rate, mean pause duration, pause word ratio, total speech duration, short/medium/long pause count, speech rate, word/audio/(filled or unfilled) pauses durations, the mean/variance/kurtosis/skewness of the first 42 Mel Frequency Cepestral Coefficients
    \item Syntactic-semantic features: probabilistic context-free grammar parsing tree heights (average / max / etc.), and the occurrences of 104 production rules (e.g: NP $\rightarrow$ DT).
    \item Lexical and POS-derived features: the occurrences of part-of-speech tags, Brunet's index, Honore's statistics, word length, cosine distances between words in sentences, etc.
\end{enumerate}

\end{document}